\pdfoutput=1

\documentclass[11pt]{article}

\usepackage[final]{acl}

\usepackage{times}
\usepackage{latexsym}

\usepackage[T1]{fontenc}

\usepackage[utf8]{inputenc}

\usepackage{microtype}

\usepackage{inconsolata}

\usepackage{graphicx}

\usepackage{booktabs}
\usepackage{subcaption}
\usepackage{listings}
\newcommand{\fumiyau}[1]{{#1}}

\title{Which Programming Language and What Features at Pre-training Stage \\ Affect Downstream Logical Inference Performance?}

\author{
 \textbf{Fumiya Uchiyama},
 \textbf{Takeshi Kojima},
 \textbf{Andrew Gambardella}, \\
 \textbf{Qi Cao},
 \textbf{Yusuke Iwasawa},
 \textbf{Yutaka Matsuo}
\\
 The University of Tokyo, Japan
\\
\href{mailto:uchiyama-fumiya014@g.ecc.u-tokyo.ac.jp}{uchiyama-fumiya014@g.ecc.u-tokyo.ac.jp}
}

\begin{document}
\maketitle
\begin{abstract}
Recent large language models (LLMs) have demonstrated remarkable generalization abilities in mathematics and logical reasoning tasks.
Prior research indicates that LLMs pre-trained with programming language data exhibit high mathematical and reasoning abilities; however, this causal relationship has not been rigorously tested.
Our research aims to verify which programming languages and features during pre-training affect logical inference performance.
Specifically, we pre-trained decoder-based language models from scratch using datasets from ten programming languages (e.g., Python, C, Java) and three natural language datasets (Wikipedia, Fineweb, C4) under identical conditions. Thereafter, we evaluated the trained models in a few-shot in-context learning setting on logical reasoning tasks: FLD and bAbi, which do not require commonsense or world knowledge.
The results demonstrate that nearly all models trained with programming languages consistently outperform those trained with natural languages, indicating that programming languages contain factors that elicit logic inference performance.
In addition, we found that models trained with programming languages exhibit a better ability to follow instructions compared to those trained with natural languages.
Further analysis reveals that the depth of Abstract Syntax Trees representing parsed results of programs also affects logical reasoning performance.
These findings will offer insights into the essential elements of pre-training for acquiring the foundational abilities of LLMs.\footnote{\fumiyau{Code is available at \url{https://github.com/fumiyauchiyama/code_pretraining}}}
\end{abstract}

\section{Introduction}
Recently, large language models (LLMs) have demonstrated remarkable generalization abilities in downstream tasks. These tasks include not only fundamental natural language processing tasks, such as machine translation and text classification~\citep{NEURIPS2020_1457c0d6}, as well as advanced tasks, such as mathematics and logical reasoning~\citep{achiam2023gpt}.

The generalization ability of LLMs originates from pre-training on large text corpora, such as RedPajama~\cite{together2023redpajama} and Fineweb~\cite{penedo2024finewebdatasetsdecantingweb}. The corpora often contain content from various domains, such as Common Crawl, GitHub, ArXiv, Wikipedia, and StackExchange. However, the relationship between each domain of training data and the abilities of LLMs is not fully understood.

Prior research has shown that LLMs pre-trained with programming language data acquire high mathematical and reasoning abilities~\citep{roziere2023code, madaan-etal-2022-language, liang2023holistic, li2023starcoder}; however, this causal relationship has not been rigorously tested.
Specifically, fair comparisons are often not conducted between models trained on programming language data and those trained on natural language data due to differences in the number of training tokens and model sizes, or because the information is unknown as for closed models.
In addition, 
 some prior works have fine-tuned models using a mixture of programming languages, but they have not conducted detailed analyses regarding the effect of each programming language on the performance of downstream tasks ~\cite{li2023starcoder, roziere2023code}.

We conducted experiments to analyze whether models trained solely on a single programming language generalize better to pure logical reasoning tasks compared to models trained on natural language. 
Specifically, we trained GPT2-124M, GPT2-774M, GPT2-1.5B, and LLaMA-774M~\cite{radford2019language, zhang2024tinyllama} from scratch using datasets from ten programming languages (e.g., Python, C, Java) and three natural language datasets (Wikipedia, Fineweb, C4) under the same conditions. We then evaluated each trained model in a few-shot in-context learning (ICL) setting on two logical reasoning tasks: Formal Logic Deduction (FLD)~\cite{morishita2023fld} and bAbi~\cite{weston2015towards}, which do not require commonsense or world knowledge.

Experimental results demonstrate that nearly all models trained on programming languages consistently outperform those trained on natural language in both FLD and bAbi.
These results indicate that it is not a particular programming language that affects logical inference performance; rather programming languages as a whole contain factors that elicit logical inference capabilities.

We qualitatively analyzed the output of each trained model and found that models trained with programming data exhibit a better ability to follow instructions compared to those trained with natural languages.
In other words, the ability to respond in the correct format, along with logical reasoning ability, is necessary, and training with programming data provides models with both abilities.
Additional experiments have confirmed that these abilities were preserved to some degree even when the commented-out parts were removed from the code.

Further analysis reveals that the complexity of code syntax, specifically the number of hierarchies, such as loops and conditional statements (e.g., "if"), also affects logical reasoning performance.
Moreover, we evaluated the trained models on the GLUE benchmark ~\cite{wang-etal-2018-glue}, and found significant differences in performance (accuracy / F1) across languages in semantic equivalence judgment tasks.

\section{Related Work}
\subsection{LLMs and Programming Language}
Two main approaches exist for solving code tasks using language models. 
One approach involves fine-tuning a model pre-trained on natural language datasets with code datasets, which is widely applied to some open models ~\citep{roziere2023code}. For closed-source models, code-davinci-002 outperforms text-davinci-002 on serialized commonsense reasoning and mathematical tasks~\cite{madaan-etal-2022-language, liang2023holistic}. 

The other approach involves training models from scratch solely on code datasets, often using a mixture of multiple programming languages. 
This method is commonly used in code completion and code generation fields. For example, SantaCoder~\cite{allal2023santacoder} is pre-trained on three mixed programming languages on The Stack~\citep{kocetkov2023the} and demonstrated superior performance not only on code completion tasks but also on the HELM benchmark ~\cite{liang2023holistic} compared to GPT-NeoX~\cite{gpt-neox-20b}.

In this study, we trained models from scratch with a single programming language under identical conditions to assess performance differences by language. In addition, we focused on measuring logical inference ability, which does not need world knowledge or common sense.

\subsection{LLMs and Logical Inference}
\label{sec:llm_logical_inf}
\citet{weston2015towards} shows that language models can solve bAbi tasks, which consist of simple logical reasoning challenges.
\citet{morishita2023fld} demonstrates that a fine-tuned T5 model can effectively address Formal Logic Deduction (FLD) tasks, involving multi-step logical reasoning.
Although these studies show that LLMs have some logical reasoning abilities, it remains unclear which features of the corpus contribute to the emergence of advanced complex reasoning.

Our study sheds light on the effects of programming languages on training LLMs.
Our findings show that the LLMs pre-trained with a single programming language outperform those trained with natural language on logical reasoning tasks. These results suggest a new criterion for corpus quality in the efficient training of LLMs.

\section{Experimental Setup}
\label{sec:setup}
\subsection{Models and Datasets}
The default model for our experiments is GPT2-small (124M). To accomodate long context few-shot in-context evaluation, we extended the model's context length from 1,024 to 2,048 tokens. \fumiyau{We employed the official GPT2 tokenizer distributed by Hugging Face \footnote{\url{https://huggingface.co/openai-community/gpt2}}} and three natural language datasets: Wikipedia~\citep{wikidump}, FineWeb~\citep{penedo2024finewebdatasetsdecantingweb}, C4~\cite{JMLR:v21:20-074}, and ten common programming languages: Haskell, OCaml, Erlang, Python, C, C++, HTML, JavaScript, TypeScript, Java from the Stack~\citep{kocetkov2023the}. 

\subsection{Evaluation Metrics}
We evaluated pre-trained models on the FLD~\cite{morishita2023fld} and bAbi~\cite{weston2015towards} datasets \fumiyau{with 3-shot ICL} using lm-evaluation-harness~\citep{abaskohi-etal-2023-lm}. 
The bAbI dataset is for simpler, more natural questions with specific word answers, while the FLD dataset involves multi-step reasoning and the specific type of answers, like `PROVED', `DISPROVED', and `UNKNOWN'.
Considering the premise and hypothesis, FLD is required to output a proof and the final answer if a hypothesis is correct based on premises. \fumiyau{However, our experiments let the models directly output the final answer without any proof because we assumed that generating natural language proof without fine-tuning is hard for small models trained on code and the lm-evaluation-harness does not support evaluating the correctness of FLD proofs.} We measured the accuracy of the final answers for both FLD and FLD* (a more complex version).

\subsection{Training Settings}
To train the language model, approximately 200M tokens were sampled from each dataset and packed each sample into fixed-length datasets, using <|$endoftext$|> tokens as delimiters. We pre-trained the models for three epochs with a batch size of 24, employing a CosineLRScheduler that warms up the learning rate linearly to 1e-4 during the first 10\% of the total iterations. The optimizer used was AdamW with $\beta_1=0.9$, $\beta_2=0.999$, $\epsilon=1e-8$, weight decay of 0.01, and gradient clipping set to 1.0. \fumiyau{We trained the models three epochs}. Other configurations are available in Appendix \ref{appx:training}.

\begin{table}
  \centering
  \begin{tabular}{lcccc}
    \hline
    \textbf{Dataset} & \textbf{FLD} & \textbf{FLD*} & \textbf{bAbi} \\
    \hline
    Wiki      & 0.14±0.00 & 0.12±0.00 & 0.01±0.00 \\
    Fineweb   & 0.00±0.00 & 0.00±0.00 & 0.00±0.00 \\
    C4   & 0.00±0.00 & 0.00±0.00 & 0.00±0.00 \\
    \hline
    Haskell   & \textbf{0.35±0.01} & \textbf{0.34±0.01} & 0.03±0.00 \\
    OCaml     & 0.32±0.01 & 0.31±0.01 & 0.05±0.00 \\
    Erlang    & 0.29±0.01 & 0.28±0.01 & 0.04±0.00 \\
    Python    & 0.34±0.01 & 0.33±0.01 & \textbf{0.07±0.00} \\
    C         & 0.34±0.01 & 0.33±0.01 & 0.06±0.00 \\
    C++       & 0.34±0.01 & 0.32±0.01 & 0.04±0.00 \\
    HTML      & 0.33±0.01 & 0.33±0.01 & 0.05±0.00 \\
    JS        & 0.33±0.01 & 0.32±0.01 & 0.03±0.00 \\
    TS        & 0.30±0.01 & 0.29±0.01 & 0.03±0.00 \\
    Java      & 0.05±0.00 & 0.06±0.00 & 0.04±0.00 \\\hline
  \end{tabular}
  \caption{Few-shot logical inference accuracy of the models pre-trained on each dataset. \fumiyau{Abbreviations used: Wiki for Wikipedia, JS for JavaScript, and TS for TypeScript. Values are presented as mean ± standard error}}
  \label{tab:lang-few}
\end{table}

\begin{figure}[t]
  \centering
  \includegraphics[width=1.00\columnwidth]{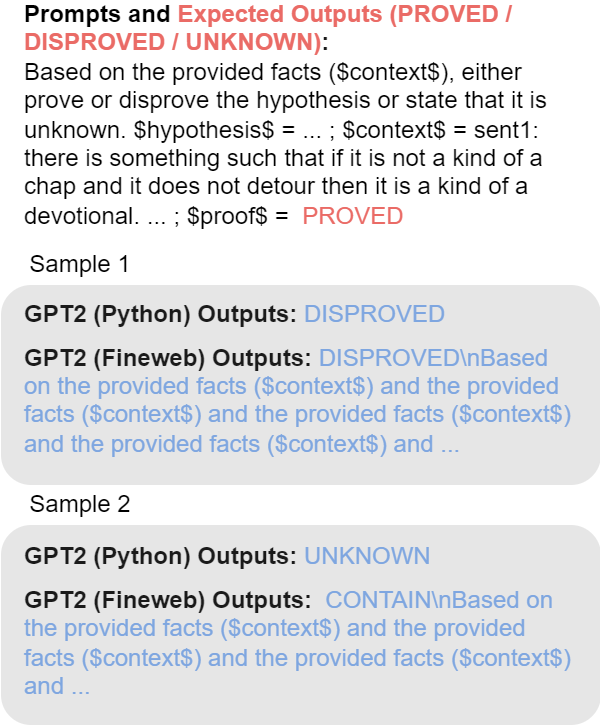}
  \caption{Sample outputs of the models trained on Python and Fineweb. Considering 3-shot examples, the model trained on Python produces a proper class name following the instruction, whereas the model trained on FineWeb produces unstructured outputs.}
  \label{fig:emnlp001sample}
\end{figure}

\section{Results}

\subsection{Logical Inference Ability by Different Programming Languages}
\label{sec:exp001}
Table~\ref{tab:lang-few} shows the accuracy of the pre-trained models with each programming language and natural language measured by FLD and bAbi datasets. 
Considering FLD and FLD*, although the best results of all models remained almost at a chance rate, the results show that the models trained in programming languages outperform the models trained in natural languages.
For bAbi, \fumiyau{code datasets influenced better performance than natural language datasets}.
Among the programming languages, Python and C showed slightly better performance across all tasks. 
However, regardless of the paradigm or typing explicitness of each language, most of the code-trained models showed better performance than natural language based models. 
The result indicates that logical inference ability and formatted outputs do not originate from a specific language but from the nature of programming itself.

Figure~\ref{fig:emnlp001sample} shows sample outputs from the models trained on Python and FineWeb.
The model trained on Python outputs in the correct format following few-shot examples, while the model trained on Fineweb outputs redundant or non-existent choice. This result is consistent with related work showing that LLMs for code have superiority over natural language based models on serialized commonsense reasoning tasks~\cite {madaan-etal-2022-language}. 

\begin{table}
  \centering
  \small
  \begin{tabular}{lcccc}
    \hline
    \textbf{Language} & \textbf{FLD} & \textbf{FLD*} & \textbf{bAbi} \\
    \hline
    Python(Shallow) & \textbf{0.35±0.01} & \textbf{0.33±0.01} & 0.05±0.00 \\
    Python(Middle) & 0.33±0.01 & \textbf{0.33±0.01} & \textbf{0.07±0.00} \\
    Python(Deep) & 0.25±0.01 & 0.25±0.01 & 0.06±0.00 \\
    \hline
  \end{tabular}
  \caption{Relationship between the complexity of programming languages and logical inference performance.} 
  \label{tab:exp004}
\end{table}

\subsection{Complexity of \fumiyau{Syntax in Code Data}}
\label{sec:exp002}
Programming languages have more complex syntax trees than natural language syntax trees and might be beneficial for reasoning in complex tasks. 
The deeper the depth of Abstract Syntax Tree (AST) — that is, the number of hierarchies consisting of elements, such as loops and conditional statements (e.g., ``if'') — the more complex the program is. 
We chose Python as the target language and separated the datasets into three subsets by AST depth: Shallow (up to 7), Middle (from 8 to 11), and Deep (from 12 to 20). \fumiyau{Each dataset is made from samples that Python ast module succeeded in parsing. Codes that did not succeed in parsing were excluded.} We trained the model on each dataset and evaluated the logical inference ability using FLD and bAbi datasets. 

Table~\ref{tab:exp004} shows the accuracy of the model trained on FLD and bAbi datasets. For bAbi, Python datasets with middle complexity show the best accuracy. For FLD, datasets of shallow complexity show the best performance, and the accuracy decreases as the depth of AST increases. 
Further investigation reveals that the model trained on the Deep dataset frequently outputs long blanks, i.e., the model outputs do not follow the instructions.
It is possible that long and complex code sentences in the training data are often indented by blanks or tabs as necessary to ensure human readability.
This redundancy in the code training data may result in the trained model outputting long blanks.
\fumiyau{In addition, we assume that there might be suitable syntax complexity to learn linguistic phenomena during pre-training.
\citet{kallini-etal-2024-mission} insists that grammar complexity of training data determines the generalization difficulty of language models for the grammar. 
}
\subsection{Ablation Study by Code Modification}
\label{sec:exp006}
\fumiyau{To further inspect what features in code raise the performance of the models on logical inference tasks, we developed three modified Python datasets: Comment-Free, Comment-Free + Scrambled, and Comment-Free + Randomized. 
"Comment-Free" is an operation that eliminates comments starting from \# and constant strings that are not used for any operation like docstring. We expected this modification to disable few-shot ICL with natural language instruction on FLD and bAbi. 
"Scrambled" shuffles identifiers (e.g. names of variables) on each position, and destroys the meaning of the code. "Randomized" replaces each identifier with a random string to cut off the knowledge of natural language. Note that syntactic correctness is maintained during all modifications. See appendix \ref{sec:code-m} for the details.

We trained models with each data on the same settings in section \ref{sec:exp001} and gained Table \ref{tab:exp006} results. 
The result shows that comment elimination maintains FLD accuracy to some extent, and cutting off learning natural languages (Comment-Free + Randomized) induces few-shot ICL failure. Destroying dependencies (Comment-Free + Scrambled) also breaks logical reasoning ability on every task. This result suggests that a language model is not a simple machine to imitate grammar, but also learns semantics from dependencies of code that can be applied to unseen logical inference tasks.}

\begin{table}
  \centering
  \small
  \begin{tabular}{lcccc}
    \hline
    \textbf{Language} & \textbf{FLD} & \textbf{FLD*} & \textbf{bAbi} \\
    \hline
    Raw & \textbf{0.34±0.01} & \textbf{0.34±0.01} & \textbf{0.05±0.00} \\
    CF & 0.23±0.01 & 0.21±0.01 & 0.04±0.00 \\
    CF+S & 0.00±0.00 & 0.00±0.00 & 0.00±0.00 \\
    CF+R & 0.00±0.00 & 0.00±0.00 & 0.01±0.00 \\
    \hline
  \end{tabular}
  \caption{Accuracy on benchmark tasks of models trained on modified code datasets. CF: Comment-Free, S: Scrambled, R: Randomized} 
  \label{tab:exp006}
\end{table}

\begin{figure*}[t]
  \centering
  \includegraphics[width=\linewidth]{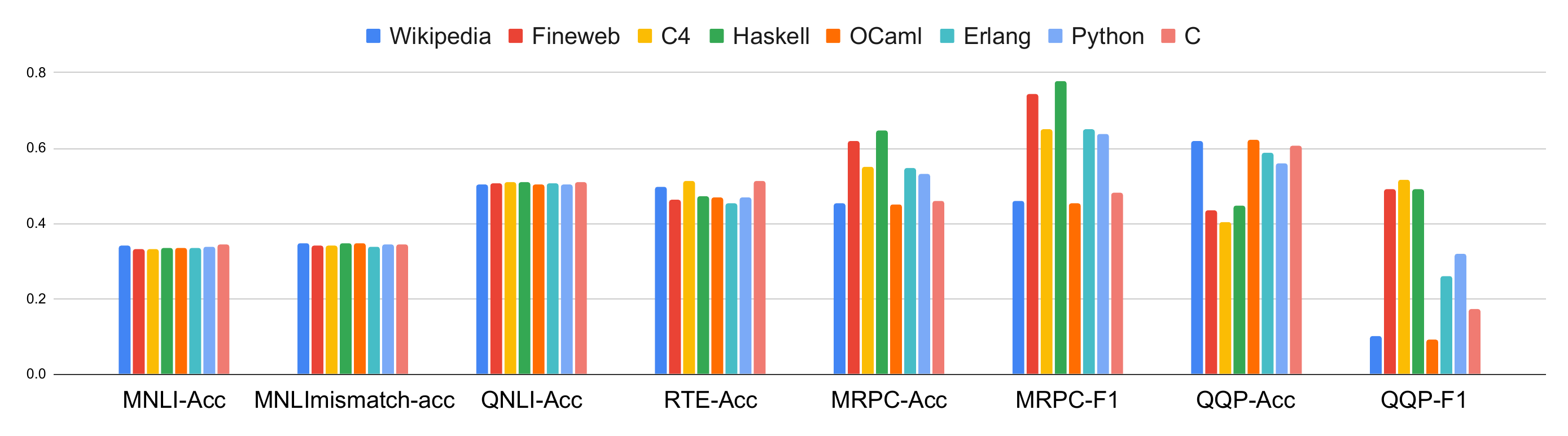}
  \vspace{-6mm}
  \caption{GLUE scores for each natural and programming language dataset. Horizontal axis represents task names and corresponding metrics (F1, Accuracy and Matthews Correlation Coefficient). Vertical axis represents task scores for each metric. Full results are available in Appendix \ref{appx:exp004}}
  \label{fig:exp004}
\end{figure*}

\subsection{Effect of Programming Language on General NLP Tasks}
\label{sec:exp004}
We also evaluated the effect of programming languages on other tasks to explore their potential as pre-training datasets.
We evaluated the pre-trained models described in Section~\ref{sec:exp001} on the GLUE benchmark ~\cite{wang-etal-2018-glue}, which focuses on natural language understanding.

\fumiyau{Figure~\ref{fig:exp004} shows the GLUE scores of the pre-trained models for each dataset. }
Entailment recognition tasks, such as MNLI, QNLI, and RTE, show that the models trained on both types of languages perform similarly. However, significant differences emerge in the performance of paraphrasing tasks, such as MRPC and QQP. Since FineWeb contains some code snippets, models trained on them may leverage specific features to enhance their understanding of semantics beyond mere syntactic differences. This is evidenced by lower F1 scores in paraphrase identification tasks like MRPC and QQP when using datasets such as Wikipedia. In contrast, datasets like C4 and FineWeb, along with certain programming languages like Haskell, achieve significantly higher scores. The presence of programming languages, even in small quantities, likely contributes positively to the models' ability to comprehend sentence meaning.

\subsection{Evaluation on Larger Models}
\label{sec:exp005}
We trained GPT2 and LLaMA, each with 774M parameters, on both Python and Fineweb. The configurations are GPT2-Large (774M) and Tiny-Llama v1.1 (1.1B)~\cite{zhang2024tinyllama}, with the MLP representation resized to 3000. \fumiyau{For LLaMA experiments, we used the official Tiny-Llama v1.1 tokenizer distributed by Hugging Face \footnote{\scriptsize{\url{https://huggingface.co/TinyLlama/TinyLlama_v1.1}}}.
For GPT2, we trained larger models based on GPT2-XL (1.5B) configuration. Specifically, 600M tokens were consumed for training GPT2-1.5B, while the other two models (GPT2-774M and LLaMA-774M) were trained on the same 200M tokens as in Section ~\ref{sec:exp001}.} 

We evaluated the models on the same tasks as described in Section~\ref{sec:exp001}. Table~\ref{tab:exp005} shows the accuracy of each programming language and natural language on FLD and bAbi. The results show that the models trained on Python outperform those trained on natural languages on FLD on both architectures. For bAbi, both models trained on Python and Fineweb show closer performance. Some scores degraded from models with 124M parameters. This is because we did not search for the best hyperparameters for model construction, and they may not have been trained under the most efficient training settings. However, the results demonstrate that code-based pre-training has superiority on logical inference ability across different model sizes and structures.

\begin{table}
  \centering
  \small
  \begin{tabular}{lcccc}
    \hline
    \textbf{Language} & \textbf{FLD} & \textbf{FLD*} & \textbf{bAbi} \\
    \hline
    GPT2(P,774M) & \textbf{0.32±0.01} & \textbf{0.32±0.01} & \textbf{0.07±0.00} \\
    GPT2(F,774M) & 0.00±0.00 & 0.00±0.00 & 0.06±0.00 \\
    \midrule
    LLaMA(P,774M) & \textbf{0.28±0.01} & \textbf{0.25±0.01} & 0.00±0.00 \\
    LLaMA(F,774M) & 0.00±0.00 & 0.00±0.00 & 0.00±0.00 \\
    \midrule
    GPT2(P,1.5B) & \textbf{0.32±0.01} & \textbf{0.31±0.01} & \textbf{0.05±0.00} \\
    GPT2(F,1.5B) & 0.00±0.00 & 0.00±0.00 & 0.04±0.00 \\
    \hline
  \end{tabular}
  \caption{Model size scale-up study. Few-shot logical inference performance of GPT2 and LLaMA with 774M \fumiyau{and 1.5B} parameters, pre-trained on each language. Abbreviations: P for Python, F for FineWeb.}

  \label{tab:exp005}
\end{table}

\section{Conclusion}
Our study rigorously verified that nearly all models trained on individual programming languages consistently achieve the better logical inference performance than those trained solely on natural language datasets \fumiyau{in few-shot ICL settings}. 
Further analysis reveals that an appropriate level of syntax complexity influences logical reasoning performance.
Additionally, models trained on programming languages exhibit a greater ability to follow instructions compared to those trained on natural language datasets. Moreover, dependencies expressed in code significantly contribute to logical reasoning in few-shot ICL settings.
We hope these findings will offer insights into the essential elements of pre-training for acquiring the foundational abilities of LLMs.

\section{Limitation}
Owing to the limitation of the computational resources, we could not train the models larger than \fumiyau{1.5B} parameters. Especially for FLD tasks, logical inference ability is limited even in models with 10 billion parameters~\citep{morishita-etal-2024-jfld}. Future work includes investigations into the effect of code-based pre-training with larger models to verify that logical reasoning abilities are more explicitly improved.
Each dataset is primarily organized in either natural language or a single programming language, although we did not conduct thorough filtering to ensure complete exclusivity. 

\fumiyau{In Section \ref{sec:exp002}, we fixed grammar complexity by selecting a single language and examined the syntax complexity in code data. However, our experiments did not consider semantic complexity or other complexities that might be measureable in both programming and natural languages. Furthermore, it remains unclear whether syntax complexity in pre-training data alone influences logical inference performance. Comparing various complexities between natural and programming language regarding logical reasoning abilities is an important avenue for future research.

In section \ref{sec:exp004}, we assessed the general language understanding of the trained models. The natural language inference tasks in GLUE require commonsense knowledge, which may be difficult to acquire through code-only training. Future experiments could explore whether fine-tuning models pre-trained on code with GLUE datasets enhances natural language reasoning capabilities. Additionally, integrating both code and natural language datasets during the pre-training process may provide a synergistic approach to leverage the strengths of both types of data.

Moreover, a further experiment in Appendix \ref{sec:ft} demonstrates the advantage on FLD tasks between natural language and programming language is reversed when fine-tuning on FLD corpus. We empathize that the advantage of logical reasoning tasks is observed in in-context learning settings and should investigate the difference between the two learning settings for logical reasoning tasks.}

\bibliography{anthology, custom}

\appendix

\section{Training Details}
\label{appx:training}
We limited training models to three epochs because several studies indicate that training language models for many epochs can worsen performance on downstream tasks and does not significantly reduce validation loss \cite{xue2024repeat, muennighoff2024scaling}.
We trained each model using a single seed for each task, resulting in a total of 26 models. Except for the experiments in Section~\ref{sec:exp005}, training took less than a day with a single NVIDIA RTX A5000 or A6000. 
For the experiments in Section~\ref{sec:exp005}, we trained each model on a single server equipped with 8 H100 GPUs for a maximum of three days.

\section{Dataset information of Section ~\ref{sec:exp002}}
\begin{figure}[ht]
  \includegraphics[width=\columnwidth]{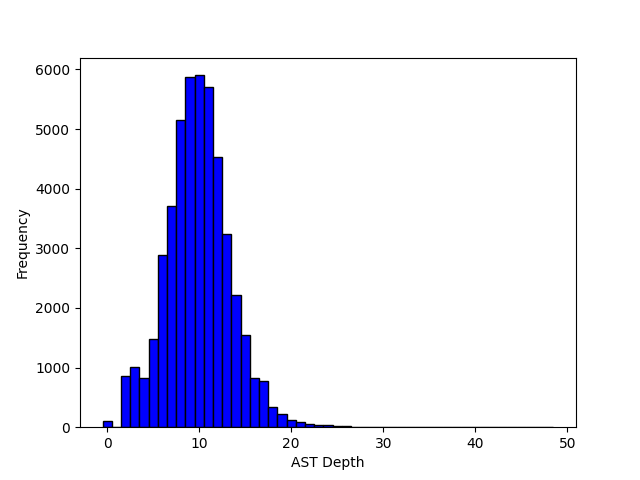}
  \caption{\fumiyau{The Frequency distribution of AST depth of 47,710 samples in The Stack Python dataset. Note that there are 14 samples whose AST depth are more than 50.}}
  \label{fig:freq-python}
\end{figure}

\fumiyau{We determined the span of AST depth for making each Python subset by referencing the distribution of AST depth in the whole dataset. Figure ~\ref{fig:freq-python} shows the histogram of 47,710 samples that are successfully parsed by Python ast module in the Stack Python 50,000 samples. Most samples have AST with depth under 20, and samples with 8-12 AST depth occupy a large portion. Therefore, we set each span of AST depth as [0,7], [8,11] and [12,20].}

\section{Logical Inference on different vocabulary}
The original FLD dataset converts formulated prompts and proofs into natural languages because it is designed for deductioning in natural languages. Although a name of a variable has some degree of freedom, a program consists of limited expression and vocabularies. Therefore, the model trained on programming languages can have a better logical inference ability when utilizing simplified vocabulary rather than complex vocabulary. 

To investigate the difference between tokens that appeared in the prompt, we evaluated the logical inference ability of the model trained during section~\ref{sec:exp001} with the formulated prompt. \fumiyau{The following example shows a default prompt and formulated prompt of FLD. Note that line breaks are added for visibility.}

\begin{quote}
\textbf{Default Prompt:} Based on the provided facts ($context$), either prove or disprove the hypothesis or state that it is unknown. \\
$hypothesis$ = the Eurasian does not occur if the hospitableness happens. ; \\
$context$ = sent1: the avoidance occurs and the Eurasian happens if the sculling does not occur. sent2: that the palpatoriness and the hospitableness occurs prevents that the Eurasian occurs. ; \\
$proof$ = 

\textbf{Formulated Prompt:} Based on the provided facts ($context$), either prove or disprove the hypothesis or state that it is unknown. \\
$hypothesis$ = ${B} \Rightarrow \neg C ;$ \\
$context$ = $sent1: \neg E \Rightarrow ({EI} \& {C})$ \\sent2: $({A} \& {B}) \Rightarrow \neg C ;$ \\
$proof$ =

\end{quote}
We chose a formulated hypothesis and context and inputted to models as a prompt. Table \ref{tab:lang-few-vocab} shows the evaluation result of models trained in Section ~\ref{sec:exp001} on the formulated prompt. The model trained on the Python dataset shows consistent performance even when the vocabulary has been changed.

\begin{table}
  \centering
  \small
  \begin{tabular}{lccc}
    \hline
    \textbf{Language} & \textbf{FLD} & \textbf{FLD*} \\
    \hline
    Fineweb & 0.00±0.00 & 0.00±0.00 \\
    Python & 0.33±0.01 & 0.33±0.01 \\
    \hline
  \end{tabular}
  \caption{Logical inference performance on the formulated prompt.}

  \label{tab:lang-few-vocab}
\end{table}

\section{Code Modification in section \ref{sec:exp006}}
\label{sec:code-m}
\fumiyau{To eliminate comments, we parsed each code sample by Python ast module. A code is transformed into AST. Subsequently, we eliminated nodes of constant strings written as statements and not used them for any assignment from AST. Because comments starting from \# is dropped when parsing code into AST, we can obtain a comment-free code by unparsing the modified AST. Scrambling and Randomizing identifiers are conducted with the same pipeline. After parsing into AST, "Scrambled" replaces names of variables, functions, classes, arguments, attributes, and imports with names sampled from a uniform distribution of all identifiers appearing in a code. "Randomized" replaces them with 8-character random strings while maintaining dependencies. Finally, we can provide a code with destroyed meanings or word knowledge of natural language by unparsing. The following listings shows a sample of each process.}


\begin{lstlisting}[caption=Raw Example]
# UCF Senior Design 2017-18
# Group 38

from PIL import Image
import cv2
import imagehash
import math
import numpy as np

DIFF_THRES = 20
LIMIT = 2
RESIZE = 1000


def calc_hash(img):
    """
    Calculate the wavelet hash of the image
        img: (ndarray) image file
    """
    # resize image if height > 1000
    img = resize(img)
    return imagehash.whash(Image.fromarray(img))


def compare(hash1, hash2):
    """
    Calculate the difference between two images
        hash1: (array) first wavelet hash
        hash2: (array) second wavelet hash
    """
    return hash1 - hash2


def limit(img, std_hash, count):
    """
    Determine whether image should be removed from image dictionary in main.py
        img: (ndarray) image file
        std_hash: (array) wavelet hash of comparison standard
        count: (int) global count of images similar to comparison standard
    """
    # calculate hash for given image
    cmp_hash = calc_hash(img)

    # compare to standard
    diff = compare(std_hash, cmp_hash)

    # image is similar to standard
    if diff <= DIFF_THRES:
        # if there are 3 similar images already, remove image
        if count >= LIMIT:
            return 'remove'

    # non-similar image found
    else:
        # update comparison standard
        return 'update_std'

    # else continue reading images with same standard
    return 'continue'


def resize(img):
    """
    Resize an image
        img: (ndarray) RGB color image
    """
    # get dimensions of image
    width = np.shape(img)[1]
    height = np.shape(img)[0]

    # if height of image is greater than 1000, resize it to 1000
    if width > RESIZE:
        # keep resize proportional
        scale = RESIZE / width
        resized_img = cv2.resize(
            img, (RESIZE, math.floor(height / scale)), cv2.INTER_AREA)
        # return resized image
        return resized_img

    # if height of image is less than 1000, return image unresized
    return img


def set_standard(images, filename):
    """
    Set new comparison standard and update information
        images: (dictionary) dictionary containing all the image data
        filename: (String) name of the image file
    """
    return filename, calc_hash(images[filename]), 0

\end{lstlisting}

\begin{lstlisting}[caption=Comment-Free Example]
from PIL import Image
import cv2
import imagehash
import math
import numpy as np
DIFF_THRES = 20
LIMIT = 2
RESIZE = 1000

def calc_hash(img):
    img = resize(img)
    return imagehash.whash(Image.fromarray(img))

def compare(hash1, hash2):
    return hash1 - hash2

def limit(img, std_hash, count):
    cmp_hash = calc_hash(img)
    diff = compare(std_hash, cmp_hash)
    if diff <= DIFF_THRES:
        if count >= LIMIT:
            return 'remove'
    else:
        return 'update_std'
    return 'continue'

def resize(img):
    width = np.shape(img)[1]
    height = np.shape(img)[0]
    if width > RESIZE:
        scale = RESIZE / width
        resized_img = cv2.resize(img, (RESIZE, math.floor(height / scale)), cv2.INTER_AREA)
        return resized_img
    return img

def set_standard(images, filename):
    return (filename, calc_hash(images[filename]), 0)

\end{lstlisting}

\begin{lstlisting}[caption=Comment-Free + Scrambled Example]
from PIL import DIFF_THRES
import img
import images
import height
import resized_img as LIMIT
RESIZE = 20
hash1 = 2
resize = 1000

def calc_hash(count):
    std_hash = std_hash(resized_img)
    return cv2.imagehash(diff.calc_hash(resized_img))

def width(img, resized_img):
    return limit - std_hash

def width(set_standard, Image, resize):
    width = height(hash1)
    height = filename(diff, RESIZE)
    if images <= compare:
        if scale >= height:
            return 'remove'
    else:
        return 'update_std'
    return 'continue'

def calc_hash(resize):
    hash2 = count.math(DIFF_THRES)[1]
    height = RESIZE.cv2(LIMIT)[0]
    if Image > hash1:
        resized_img = count / resized_img
        img = limit.resized_img(set_standard, (calc_hash, calc_hash.compare(cv2 / imagehash)), width.calc_hash)
        return Image
    return hash1

def DIFF_THRES(Image, img):
    return (limit, resize(img[DIFF_THRES]), 0)

\end{lstlisting}

\begin{lstlisting}[caption=Comment-Free + Randomized Example]
from WOLFjkmq import aCux4Y4Q
import Q1pG5gl3
import Gx1YslqS
import T3HRhbs3
import LJTWG4w8 as GCBgPcV2
Ges4set_ = 20
tm74wylu = 2
zln4AZrv = 1000

def lZ50hv90(wPSRoTdu):
    wPSRoTdu = wewPZ1Mm(wPSRoTdu)
    return Gx1YslqS.fjqin3Y_(aCux4Y4Q._am0qTs7(wPSRoTdu))

def CX7r6rrH(MSI8x6sB, M6wvOBrw):
    return MSI8x6sB - M6wvOBrw

def OwRQZArW(wPSRoTdu, aJUeLgwi, dQ0rdVnl):
    qfSknjgG = lZ50hv90(wPSRoTdu)
    SXIn4PMr = CX7r6rrH(aJUeLgwi, qfSknjgG)
    if SXIn4PMr <= Ges4set_:
        if dQ0rdVnl >= tm74wylu:
            return 'remove'
    else:
        return 'update_std'
    return 'continue'

def wewPZ1Mm(wPSRoTdu):
    ldiBeObH = GCBgPcV2.P9O5IlYb(wPSRoTdu)[1]
    XsvyluRz = GCBgPcV2.P9O5IlYb(wPSRoTdu)[0]
    if ldiBeObH > zln4AZrv:
        _017HwMd = zln4AZrv / ldiBeObH
        zShzC25m = Q1pG5gl3.wewPZ1Mm(wPSRoTdu, (zln4AZrv, T3HRhbs3.F2fRx57k(XsvyluRz / _017HwMd)), Q1pG5gl3.pI7RGMeM)
        return zShzC25m
    return wPSRoTdu

def TgNnQBZK(Qd_fVhjP, tqVDS33U):
    return (tqVDS33U, lZ50hv90(Qd_fVhjP[tqVDS33U]), 0)
\end{lstlisting}

\section{Fine-tuning on FLD corpus}
\label{sec:ft}
\fumiyau{We have demonstrated the reasoning skills of LLMs in a few-shot in-context learning setting. However, fine-tuning is another method to achieve domain specialization. Because the answer accuracy of FLD in in-context learning is almost the same as the chance rate, we fine-tuned models trained in section \ref{sec:exp005}. We utilized the official fine-tuning code provided by \citealp{morishita-etal-2024-jfld}. During the training models generate proofs and the final answers as supervised learning while the other experiments let models output the final answers directly. In particular, 10,000 samples are used as a training data, and 500 samples are used for evaluation data. 

Table \ref{tab:ft-fld} shows the answer accuracy on FLD evaluation of models fine-tuned on FLD corpus. The model pre-trained on Fineweb outperforms the other model pre-trained on Python. This result is contrary to that in in-context learning settings, and implies that different datasets are suited for improving in-context learning ability for unseen tasks versus domain specialization ability for logical reasoning.}

\begin{table}
  \centering
  \small
  \scalebox{0.92}[1.0]{
  \begin{tabular}{lcccccc}
    \hline
    \textbf{Language} & \textbf{D-0} & \textbf{D-1} & \textbf{D-2} & \textbf{D-3} & \textbf{D-None} & \textbf{D-All} \\
    \hline
    Python & 0.50 & 0.53 & 0.33 & 0.39 & 0.17 & 0.34 \\
    Fineweb & 0.42 & 0.77 & 0.64 & 0.54 & 0.17 & 0.50 \\
    \hline
  \end{tabular}
  }
  \caption{Answer accuracy on FLD evaluation of fine-tuned models. D-* means the subset of FLD separated by the depth of the proof tree. Note that each subset has a different size of samples, and D-None is a subset composed of unprovable problems then there are no proofs. D-All is the accuracy of the whole FLD evaluation.}

  \label{tab:ft-fld}
\end{table}

\section{Detailed Result of GLUE Evaluation in Section \ref{sec:exp004}}
\label{appx:exp004}

\begin{figure*}[t]
  \begin{subfigure}{\linewidth}
    \centering
    \includegraphics[width=\linewidth]{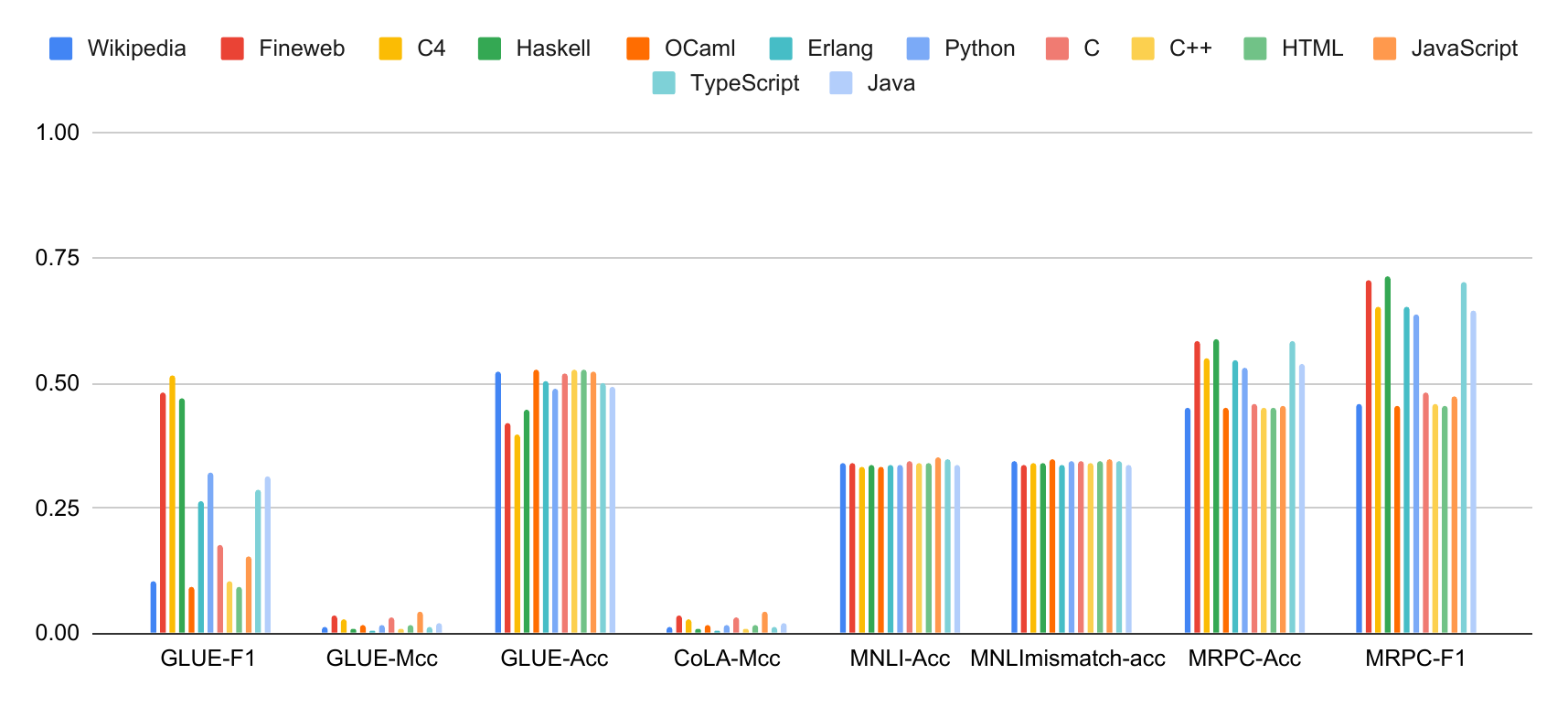}
  \end{subfigure}
  
  \begin{subfigure}{\linewidth}
    \centering
    \includegraphics[width=\linewidth]{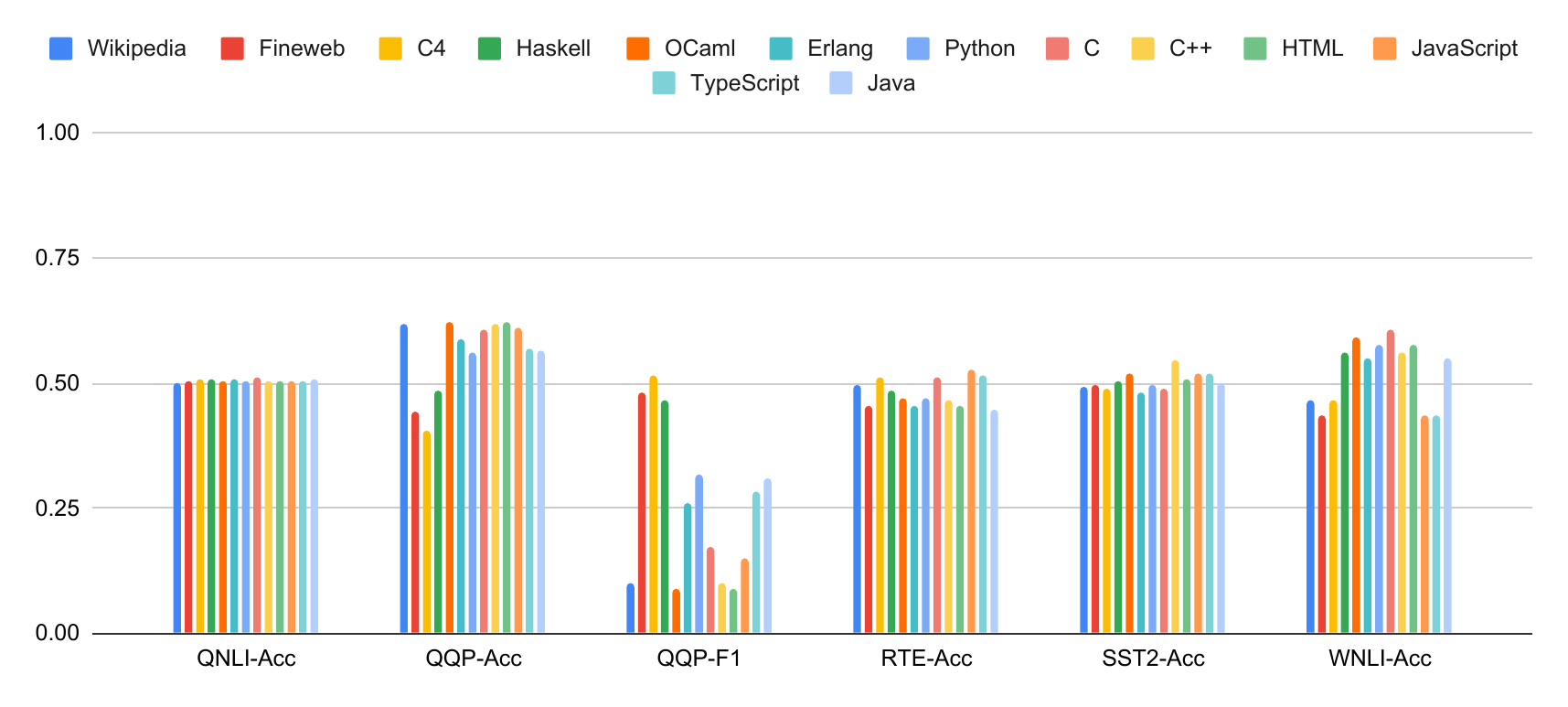}
  \end{subfigure}
  \caption{Values of GLUE score by each natural language and programming language. Horizontal axis represents the task name and its metrics (F1, Accuracy and Matthews Correlation Coefficient). Vertical axis represents the score of the task by each metrics.}
  \label{fig:exp004-ap}
\end{figure*}

Figure~\ref{fig:exp004-ap} shows the full GLUE score of pre-trained models with each of programming language and natural language datasets.

\section{License}

\vspace{-1mm}

\subsection{Model}
\begin{itemize}
    \small
    \setlength{\itemsep}{-2mm}
    \item GPT2: MIT [\href{https://huggingface.co/datasets/choosealicense/licenses/blob/main/markdown/mit.md}{link}]
    \item TinyLlama: Apache 2.0 [\href{https://huggingface.co/datasets/choosealicense/licenses/blob/main/markdown/apache-2.0.md}{link}]
\end{itemize}

\vspace{-5mm}

\subsection{Dataset}
\begin{itemize}
    \small
    \setlength{\itemsep}{-2mm}
    \item Wikipepia: cc-by-sa-3.0 [\href{https://huggingface.co/datasets/legacy-datasets/wikipedia}{link}]
    \item Fineweb: odc-by [\href{https://huggingface.co/datasets/HuggingFaceFW/fineweb}{link}]
    \item C4: odc-by [\href{https://huggingface.co/datasets/allenai/c4}{link}]
    \item The Stack: various (differed by datapoints) [\href{https://huggingface.co/datasets/bigcode/the-stack/blob/main/licenses.json}{link}]
    \item FLD: Apache 2.0 [\href{https://github.com/hitachi-nlp/FLD}{link}]
    \item bAbi: BSD License [\href{https://github.com/facebookarchive/bAbI-tasks?tab=License-1-ov-file#readme}{link}]
    \item GLUE: MIT(CoLA), OANC/CC BY-SA 3.0/CC BY 3.0(MNLI), CC BY-SA 4.0 (QNLI), MIT(QQP, SST2), Unknown(MRPC, RTE, WNLI) [\href{https://gluebenchmark.com/faq}{link}]
\end{itemize}


\end{document}